\pdfoutput=1

\documentclass[11pt]{article}

\usepackage{acl}

\usepackage{graphicx,xcolor}
\usepackage{url}
\usepackage{array}
\usepackage{here}
\usepackage{float}
\usepackage{multicol}
\usepackage{times}
\usepackage{latexsym}

\usepackage[T1]{fontenc}

\usepackage[utf8]{inputenc}

\usepackage{microtype}

\usepackage{inconsolata}

\usepackage{multirow,tabularx}
\usepackage{array}
\usepackage{here}
\usepackage{float}
\usepackage{multicol}
\usepackage{arydshln}
\usepackage {diagbox}

\usepackage{amsmath}
\usepackage{amssymb}
\usepackage{amsfonts}
\usepackage{latexsym}

\newcommand{\tabH}{\rule{0pt}{2.1ex}}

\newcommand{\footlink}[1]{\footnote{\url{#1}}}

\title{Sentence Representations via Gaussian Embedding}

\author{
  Shohei Yoda \hspace{2em} Hayato Tsukagoshi \hspace{2em} Ryohei Sasano \hspace{2em} Koichi Takeda \\
  Graduate School of Informatics, Nagoya University\\
  \texttt{yoda.shohei.a1@s.mail.nagoya-u.ac.jp}, \\
  \texttt{tsukagoshi.hayato.r2@s.mail.nagoya-u.ac.jp}, \\
  \texttt{\{sasano,takedasu\}@i.nagoya-u.ac.jp} \\
}

\begin{document}
\maketitle

\begin{abstract}
Recent progress in sentence embedding, which represents a sentence's meaning as a point in a vector space, has achieved high performance on several tasks such as the semantic textual similarity (STS) task.
However, a sentence representation cannot adequately express the diverse information that sentences contain: for example, such representations cannot naturally handle asymmetric relationships between sentences.
This paper proposes GaussCSE, a Gaussian-distribution-based contrastive learning framework for sentence embedding that can handle asymmetric inter-sentential relations, as well as a similarity measure for identifying entailment relations.
Our experiments show that GaussCSE achieves performance comparable to that of previous methods on natural language inference (NLI) tasks, and that it can estimate the direction of entailment relations, which is difficult with point representations.
\end{abstract}

\section{Introduction}
Sentence embeddings are representations to describe a sentence's meaning and are widely used in natural language tasks such as document classification~\cite{classification}, sentence retrieval~\cite{retrieval}, and question answering~\cite{qa}. 
In recent years, machine-learning-based sentence embedding methods with pre-trained language models have become mainstream, and various methods for learning sentence embeddings have been proposed~\cite{SBERT,SimCSE}.
However, as these methods represent a sentence as a point in a vector space and primarily use symmetric measures such as the cosine similarity to measure the similarity between sentences, they cannot capture asymmetric relationships between two sentences, such as entailment and hierarchical relations.

\begin{figure}[t]
\centering
\includegraphics[width=\linewidth]{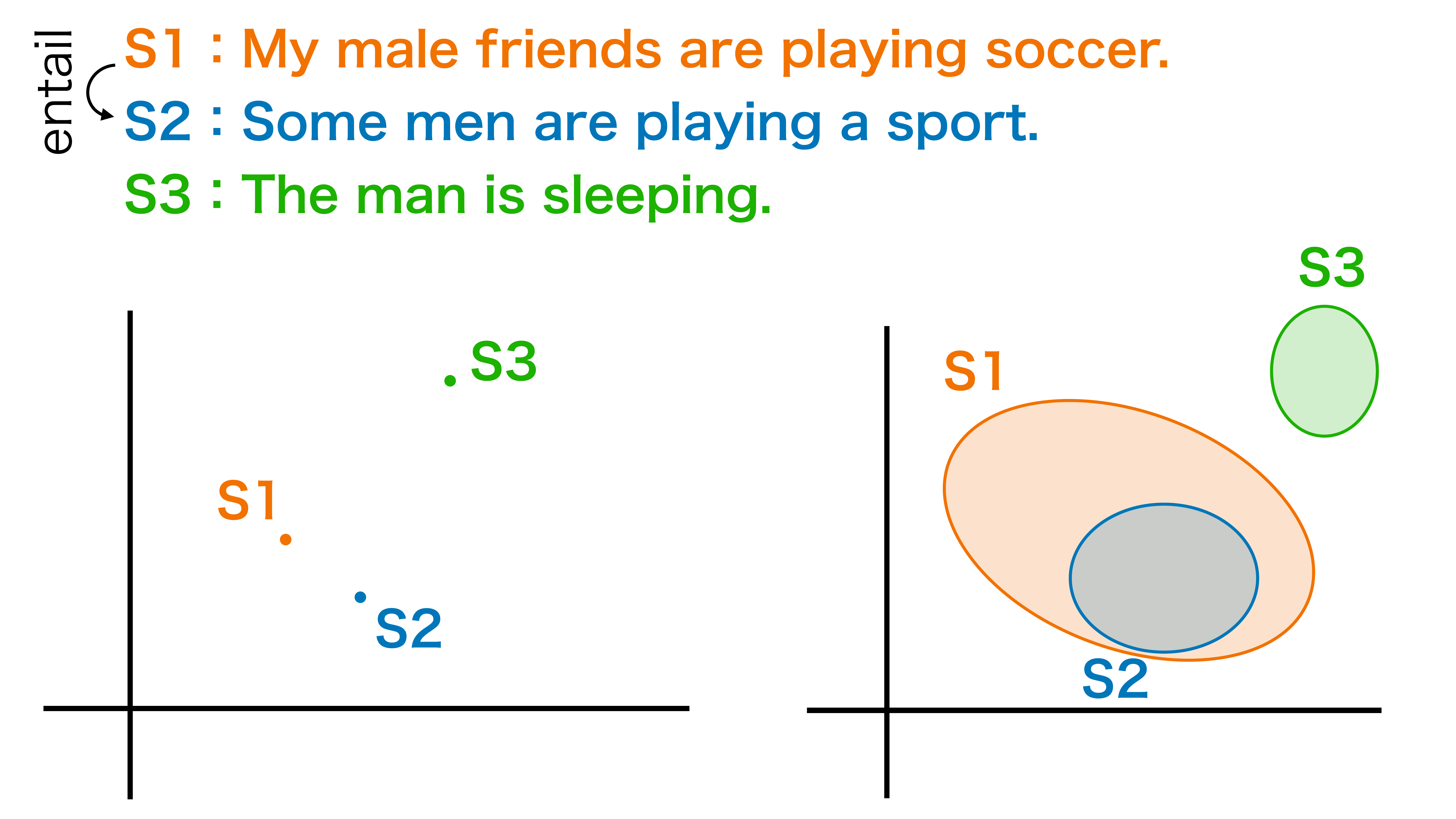}
\caption{Sentence representations in embedding spaces of a previous method (left) and GaussCSE (right).}
\label{fig:emb_space}
\end{figure}

In this paper, we propose GaussCSE, a Gaussian-distribution-based contrastive sentence embedding to handle such asymmetric relationships between sentences by extending Gaussian embedding for words \cite{Gemb}.
Figure~\ref{fig:emb_space} shows examples of sentence representations obtained by a previous method and by GaussCSE.
Whereas the previous method represents a sentence as a point, GaussCSE represents a sentence as a region in the embedding space, and when two sentences have an entailment relation, the embedding of the entailing sentence contains the embedding of the entailed one.
In these examples, S1 entails S2, but with previous methods, it is difficult to determine the entailment relation only from their embeddings.
In contrast, by taking into account the variances of the distributions, GaussCSE can capture the asymmetric relationship where S1 entails S2 but S2 does not entail S1, as well as the fact that S3 is not in the entailment relationship with either S1 or S2.

To validate the usefulness of GaussCSE, we performed comparative experiments on two tasks: the natural language inference (NLI) task, and the task of predicting the entailment direction.
The results demonstrate that GaussCSE can accurately predict the entailment direction while maintaining good performance on the NLI task.\footnote{We released our code and fine-tuned models at \url{https://github.com/yoda122/GaussCSE}.}

\section{Sentence Representations via Gaussian Embedding}
GaussCSE is a method to obtain Gaussian embeddings of sentences by fine-tuning a pre-trained language model through contrastive learning. 
In this section, we first review a representative study of Gaussian embeddings and then review SimCSE, a method that acquires sentence embeddings via contrastive learning. 
We also review embedding methods that focus on asymmetric relations, which is closely related to our research.
We then describe GaussCSE, which extends Gaussian embeddings and SimCSE.

\subsection{Gaussian Embedding}
One representative study on Gaussian embeddings sought to embed a word as a Gaussian distribution $\mathcal{N}$~\cite{Gemb}.
In this method, the embedding $N_i$ of a word $w_i$ is represented as $\mathcal{N}(x;\mu_{i},\Sigma_{i})$ by using the mean vector $\mu_i$ in $n$-dimensional space and the variance-covariance matrix $\Sigma_i$.

The similarity between two words is measured using the Kullback-Leibler (KL)  divergence, as defined by the following equation:
\begin{align}
D_{\mathrm{KL}}(N_i||N_j)&=\notag\\
\int_{x\in\mathbb{R}^n} \mathcal{N}&(x;\mu_{i},\Sigma_{i})\log\frac{\mathcal{N}(x;\mu_{i},\Sigma_{i})}{\mathcal{N}(x;\mu_{j},\Sigma_{j})}.
\end{align}

The KL divergence is an asymmetric measure whose value changes when the arguments are reversed, which makes it suitable for capturing asymmetric relationships between embeddings, such as entailment relations.

\subsection{Supervised SimCSE}
\label{sec:SimCSE}

In recent years, there has been a significant amount of research on methods for acquiring vector-based sentence embeddings~\citep[e.g.,][]{IS-BERT,BERT-flow,defsent,PromptBERT,DiffCSE,SCD}.
One of the most representative methods is supervised SimCSE~\cite{SimCSE}, which trains sentence embedding models through contrastive learning on NLI datasets. 

NLI datasets contain collections of sentence pairs, where each pair comprises a premise and a hypothesis and is labeled with ``entailment,'' ``neutral,'' or ``contradiction.'' Specifically, supervised SimCSE uses sentence pairs labeled with ``entailment'' as positive examples and those labeled with ``contradiction'' as hard negative examples. 
This approach achieves high performance on semantic textual similarity (STS) tasks, which evaluate how well sentence embedding models capture the semantic similarities between the sentences in a pair.

\subsection{Sentence Embeddings for Asymmetric Relations}

Similar to our approach, there are several studies that focus on the asymmetric relationships between sentences.
Sen2Pro~\cite{sen2pro} represents sentences as probability distributions by sampling embeddings multiple times from pre-trained language models to reflect model and data uncertainty.
RSE~\cite{rse} enriches sentence embeddings by incorporating relationships between sentences, such as entailment and paraphrasing, allowing for a more comprehensive representation of information.
Unlike these methods, we propose a fine-tuning method utilizing contrastive learning for generating probabilistic distributed representations of sentences.

\subsection{GaussCSE}
To handle asymmetric relationships between sentences, we fine-tune pre-trained language models for representing sentences as Gaussian distributions via contrastive learning.
We call this approach GaussCSE.
First, a sentence $s_k$ is fed to BERT, and the sentence's vector representation $v_k$ is obtained from the embedding of the \verb|[CLS]| token. 
When using RoBERTa, where the \verb|[CLS]| token does not exist, the beginning-of-sentence token \verb|<s>| is used as an alternative. 
Then, $v_k$ is fed to two distinct linear layers, thus obtaining a mean vector $\mu_k$ and a variance vector $\sigma_k$, which is a diagonal component of a variance-covariance matrix.
Note that, for computational efficiency, we adopt the same approach as in the previous study~\cite{Gemb}; that is, we represent the variance by using only the diagonal elements of the variance-covariance matrix.
Subsequently, we use $\mu_k$ and $\sigma_k$ to obtain a Gaussian distribution $N_k$ as a sentence representation.

We then define a similarity measure by the following equation to measure the asymmetric similarity of sentence $s_i$ with respect to sentence $s_j$:
\begin{equation}
\mbox{sim}(s_i||s_j)=\frac{1}{1+D_\mathrm{KL}(N_i||N_j)}.
\label{eq:sim}
\end{equation}
Because the KL divergence's range is $[0, \infty)$, the range of $\mbox{sim}(s_i||s_j)$ is $(0, 1]$.
When the variance of $N_i$ is greater than the variance of $N_j$, $D_\mathrm{KL}(N_i||N_j)$ tends to be larger than $D_\mathrm{KL}(N_j||N_i)$, which means that $\mbox{sim}(s_j||s_i)$ tends to be larger than $\mbox{sim}(s_i||s_j)$.
Note that $\mbox{sim}(s_j||s_i)$ can be computed with the same computational complexity as cosine similarity, owing to representing the variance using only the diagonal elements of the variance-covariance matrix.\footnote{More details are provided in Appendix~\ref{sec:complexity}.}

When learning entailment relations, as with word representation by Gaussian embedding, GaussCSE performs learning such that the embedding of a sentence that entails another sentence has greater variance than the embedding of the sentence that is entailed.
To achieve this, we use sentence pairs in an entailment relationship and increase the variance for premise ($\textit{pre}$) sentences while decreasing it for hypothesis ($\textit{hyp}$) sentences in NLI datasets. 
This is accomplished by training the model to increase $\mbox{sim}(\textit{hyp}||\textit{pre})$ relative to $\mbox{sim}(\textit{pre}||\textit{hyp})$ in accordance with the characteristics of the KL divergence as described above. 
Conversely, we decrease $\mbox{sim}(\textit{hyp}||\textit{pre})$ when the premise does not entail the hypothesis, thus indicating that the sentences are not semantically related. 
As the KL divergence is more sensitive to differences in the mean than differences in the variance, this operation is expected to increase the distance between the two sentences' distributions.

Following the supervised SimCSE approach, we use contrastive learning with NLI datasets to train the model.
During training, we aim to increase the similarity between positive examples and decrease the similarity between negative examples. 
We use the following three sets for positive and negative examples.

\begin{description}
\item[Entailment set]{
The set of premise and hypothesis pairs labeled with ``entailment.''
These semantically similar sentences are brought closer to each other.
}
\item[Contradiction set]{
The set of premise and hypothesis pairs labeled with ``contradiction.'' 
These sentences with no entailment are used as negative examples and are spread apart from each other.
}
\item[Reversed set]{
The set of sentence pairs obtained by reversing each pair in the ``entailment set.'' 
These sentences, whose entailment relation is reversed, are used as negative examples to increase the variance of premise sentences and decrease the variance of hypothesis sentences.
}
\end{description}

We compute $\mbox{sim}(\textit{hyp}||\textit{pre})$ for both positive and negative examples. 
Specifically, the similarities of positive and negative examples in the three sets are computed by using $n$ triplets of sentences ($s_i$, $s_i^+$, $s_i^-$), where $s_i$ is premise, $s_i^+$ and $s_i^-$ are entailment and contradiction hypotheses.
The loss function for contrastive learning is defined as follows:
\begin{gather}
\label{eq:loss}
    V_{E}=\Sigma_{j=1}^{n}e^{\mathrm{sim}(s_j^+||s_i)/\tau},\notag\\
    V_{C}=\Sigma_{j=1}^{n}e^{\mathrm{sim}(s_j^-||s_i)/\tau},\notag\\
    V_{R}=\Sigma_{j=1}^{n}e^{\mathrm{sim}(s_j||s_i^+)/\tau},\notag\\
\mathcal{L}=\sum_{i=1}^{n}{-\log\frac{e^{\mathrm{sim}(s_i^+||s_i)/\tau}}{V_{E}+V_{C}+V_{R}}},
\end{gather}
where $n$ is a batch size and $\tau$ is a temperature hyperparameter.

By performing learning with such a loss function, the model is expected to learn close mean vectors for sentences that are semantically similar. 
For entailment pairs, it is expected that the variance of the entailing sentence will become large and that of the entailed sentence will become small.

\section{Experiments}
We validated the effectiveness of GaussCSE through experiments on two tasks: NLI and prediction of the entailment direction.

\subsection{NLI Task}
We evaluated GaussCSE by comparing it with previous methods for recognizing textual entailment.
NLI tasks usually perform three-way classification, but we performed two-way classification by collapsing the ``neutral'' and ``contradiction'' cases as ``non-entailment,'' following revious studies on sentence embeddings.
When the value of $\mbox{sim}(\textit{hyp}||\textit{pre})$ was greater than a threshold, the relation was classified as ``entailment''; otherwise, it was classified as ``non-entailment.''

We used the Stanford NLI (SNLI)~\cite{SNLI}, Multi-Genre NLI (MNLI)~\cite{MNLI}, and SICK~\cite{SICK} datasets for evaluation.\footnote{The details of each dataset are in Appendix~\ref{sec:dataset}}
We adopted the accuracy as the evaluation metric and we used the threshold that achieved the highest accuracy on the development set to calculate the accuracy.

\subsection{Entailment Direction Prediction Task}
To validate that GaussCSE can capture asymmetric relationships, we performed the task of predicting which sentence entailed the other when given two sentences $A$ and $B$ in an entailment relation.
We used the similarity to determine the entailment direction, where $A$ is determined to be the entailing sentence if $\mbox{sim}(\textit{B}||\textit{A})$ was larger than $\mbox{sim}(\textit{A}||\textit{B})$. 
For this task, we used only sentence pairs labeled ``entailment'' in the datasets, and we adopted the accuracy as the evaluation metric. 
Note that SICK has instances with the bilateral entailment label.
As there is no unique entailment direction between a pair of such sentences, we excluded such sentence pairs from the dataset in this experiment.

\subsection{Experimental Setup}
\label{sec:setup}
We used BERT-base, BERT-large, RoBERTa-base, and RoBERTa-large in transformers\footlink{https://github.com/huggingface/transformers} as pre-trained language models, and report the results for BERT-base and RoBERTa-large in Section~\ref{sec:result}.\footnote{All the experimental results are in Appendix~\ref{sec:nli_full} and ~\ref{sec:ent_full}.}
Following \citet{SimCSE}, we combined the SNLI and MNLI datasets to form the training dataset. 
We conducted a statistical test for differences in accuracies when using the same pre-trained language model and dataset.
Specifically, we tested the differences in accuracies obtained by the different loss functions with McNemar's test at a significance level of 0.05.
Each experiment was conducted with five different random seeds, and the average was used as the final score.
Details of other configurations are provided in the Appendix~\ref{sec:detail}.
 
We conducted experiments with four different loss functions, each with different training data: the entailment set alone (ent), the entailment and contradiction sets (ent+con), the entailment and reversed sets (ent+rev), and all sets (ent+con+rev). 

\subsection{Results}
\label{sec:result}

\begin{table}[t!]
\centering
\small
\begin{tabular}{@{ \ }l@{ \ }l|c@{ \ \ }c@{ \ \ }c|c@{ \ }}
\hline
\tabH Model & Loss function & SNLI & MNLI & SICK & Avg.\\
\hline
\multicolumn{2}{@{ }l|}{\tabH SimCSE {\small (BERT-base)}} & 74.96 & 78.18 & 86.11 & 79.75\\
\hline
          & \tabH ent & 72.44 & 67.92 & 67.70 & 69.35\\
    BERT & ent+con & \textbf{77.63} & \textbf{77.71} & 80.38 & 78.57\\
    -base & ent+rev & 69.32 & 66.04 & 67.93 & 67.76\\
          & ent+con+rev & 76.64 & 76.85 & \textbf{83.15} & \textbf{78.88}\\
\hline
            & \tabH ent & 72.54 & 68.67 & 69.96 & 70.39\\
    RoBERTa & ent+con & \textbf{78.05} & \textbf{79.96} & 81.05 & 79.68\\
    -large  & ent+rev & 69.17 & 66.47 & 67.84 & 67.82\\
            & ent+con+rev & 76.68 & 79.07 & \textbf{84.17} & \textbf{79.97}\\
\hline
\end{tabular}
\caption{Experimental results of the NLI task.}
\label{tab:nli}
\end{table}

\paragraph{NLI task}
Table~\ref{tab:nli} lists the experimental results of the NLI task. 
The performance of supervised SimCSE\footnote{\url{https://github.com/princeton-nlp/SimCSE}} trained on BERT-base is given as a baseline.
Among the four settings, those using both the entailment and contradiction sets (ent+con and ent+con+rev) performed relatively well, achieving comparable performance to that of SimCSE.
Because the reversed set comprised semantically similar sentence pairs, treating such similar sentence pairs as negative examples did not contribute to performance in the NLI task.

\paragraph{Entailment Direction Prediction Task}
Table~\ref{tab:ent} lists the experimental results of entailment direction prediction. 
The performance of a baseline method which determines longer sentence as entailing one (length-baseline) is also given.
We can see that the leveraging of the reversed set significantly improved the accuracy, and outperformed the baseline method. 
This indicates that GaussCSE succeeds in acquiring embeddings that can recognize the direction of the entailment by using the reverse set as negative examples.

\begin{table}[t]
\centering
\small
\begin{tabular}{@{ \ }l@{ \ }l|c@{ \ \ }c@{ \ \ }c|c@{ \ }}
\hline
\tabH Model & Loss function & SNLI & MNLI & SICK & Avg.\\
\hline
\multicolumn{2}{@{ }l|}{\tabH Length-baseline} & 92.63 & 82.64 & 69.14 & 81.47\\
\hline
          &\tabH ent & 64.84 & 61.11 & 60.10 & 62.01\\
    BERT  & ent+con & 64.55 & 56.84 & 69.67 & 63.68\\
    -base & ent+rev & \textbf{97.60} & \textbf{92.64} & \textbf{87.80} & \textbf{92.68}\\
          & ent+con+rev & \textbf{97.38} & \textbf{91.92} & \textbf{86.22} & 91.84\\
\hline
            & \tabH ent & 66.91 & 60.88 & 61.56 & 63.11\\
    RoBERTa & ent+con & 64.57 & 55.31 & 71.38 & 63.75\\
    -large  & ent+rev & \textbf{97.89} & \textbf{93.97} & \textbf{88.71} & \textbf{93.52}\\
            & ent+con+rev & \textbf{97.42} & 93.61 & 86.57 & 92.53\\
\hline
\end{tabular}
\caption{Experimental results of the entailment direction prediction task.
}
\label{tab:ent}
\end{table}

Regarding the differences in accuracy among the datasets, accuracies of over 97\% and over 93\% were achieved on the SNLI and MNLI datasets, respectively, whereas the accuracy on the SICK dataset was relatively low, 89\% at the highest. 
These results were presumably due to the datasets' characteristics regarding the different lengths of sentence pairs.\footnote{Sentence length ratios of these datasets are provided in Appendix~\ref{sec:ratio}.}
However, the fact that GaussCSE achieved 89\% accuracy by leveraging the reversed set even on the SICK dataset indicates that it took the semantic content of sentences into account in capturing entailment relationships.

Considering the overall experimental results of the two tasks, we can conclude that by leveraging both contradiction and reverse sets as negative examples, GaussCSE could achieve high accuracy in predicting the direction of entailment relations while retaining the performance of the NLI task.

\section{Conclusion}
In this paper, we have presented GaussCSE, a Gaussian-distribution-based contrastive sentence embedding to handle asymmetric relationships between sentences. 
GaussCSE fine-tunes pre-trained language models via contrastive learning with asymmetric similarity.
Through experiments on the NLI task and entailment direction prediction, we have demonstrated that GaussCSE achieves comparative performance to previous methods on NLI task and also accurately estimate the direction of entailment relations, which is difficult with conventional sentence representations.

In this study, we used a Gaussian distribution to represent the spread of the meaning of a sentence in the embedding space, we would like to conduct a comparison with the use of other types of embedding, such as Hyperbolic Embeddings~\cite{poincare} or Box Embeddings~\cite{word2box} in future work.

\section*{Limitations}

Our proposed method involves supervised learning to acquire Gaussian-based sentence representations, but the optimal choices of the probability distribution and domain representation are not yet known. 
Additionally, for low-resource languages on which large-scale NLI datasets may not be available for use as supervised training data, alternative training approaches will need to be explored. 
To address these challenges, future investigations could consider alternative embedding methods such as box embeddings going beyond Gaussian-based approaches, as well as experiments using multilingual models. 
Furthermore, it would be beneficial to explore unsupervised learning techniques that are less dependent on language resources.

\section*{Acknowledgements}
This work was partly supported by JSPS KAKENHI Grant Number 21H04901.

\bibliography{anthology,custom}

\appendix

\section{Computation Complexity of KL divergence}
\label{sec:complexity}
The KL divergence between Gaussian distributions can be computed analytically using the following formula:
\begin{align}
\begin{split}
D_{KL}(N_i\|N_j) &= \notag\\
\frac{1}{2}[\log \frac{|\Sigma_j|}{|\Sigma_i|} &+ tr(\Sigma_j^{-1}\Sigma_i) + \\
&({\mu_i} - {\mu_j})^T \Sigma_j^{-1} ({\mu_i} - {\mu_j}) -d],
\end{split}
\end{align}

\noindent where $d$ denotes the dimension of $N_1$ and $N_2$. 
Since we set all elements except the diagonal components of the covariance matrix to zero, $\Sigma^{-1}$ becomes the reciprocal of each component in $\Sigma$ and $|\Sigma|$ can be computed as the product of its diagonal components.
The calculations for each term can be done in $O(d)$, resulting in an overall computational complexity of $O(d)$, which is the same with the computational complexity of cosine similarity.

\section{Details of NLI Datasets}
\label{sec:dataset}
SNLI, MNLI and SICK datasets comprise pairs of premise and hypothesis sentences. 
SNLI contains approximately 570,000 sentence pairs, where the premise sentences were obtained by crawling image descriptions, and the hypothesis sentences were manually generated and annotated by human annotators.
MNLI contains approximately 430,000 sentence pairs, and its construction method was similar to that of SNLI. 
The key difference is that MNLI includes premise sentences from both written and spoken speech in a wider range of styles, degrees of formality, and topics as compared to SNLI. 
SICK contains approximately 10,000 sentence pairs. 
Like SNLI, the premise sentences in SICK were constructed from sources such as image descriptions; however, a portion of the premise sentences was automatically replaced by using specific rules to generate the hypothesis sentences.

\section{Full Results of the NLI Task}
\label{sec:nli_full}

\begin{table*}[t!]
\centering
\small
\begin{tabular}{@{ \ }l@{ \ }l|c@{ \ \ }c@{ \ \ }c@{ \ \ }c@{ \ \ }c
@{ \ \ }c}
\hline
\tabH \multirow{2}{*}{Model} & \multirow{2}{*}{Loss function}  & \multicolumn{2}{c}{SNLI} & \multicolumn{2}{c}{MNLI} & \multicolumn{2}{c}{SICK}\\
\   & \  & Acc. & AUPRC & Acc. & AUPRC & Acc. & AUPRC\\
\hline
\multicolumn{2}{l|}{\tabH SimCSE {\small (BERT-base)}} & 74.96 & 66.76 & 78.18 & 75.88 & 86.11 & 81.41\\
\hline
          & \tabH ent & 72.44 & 60.65 & 67.92 & 56.96 & 67.70 & 68.26 \\
      BERT& ent+con & \textbf{77.63} & \textbf{70.95} & \textbf{77.71} & \textbf{74.21} & 80.38 & \textbf{82.12} \\
     -base& ent+rev & 69.32 & 54.21 & 66.04 & 53.87 & 67.93 & 63.60 \\
          & ent+con+rev & 76.64 & 67.07 & 76.85 & 71.34 & \textbf{83.15} & 79.45 \\
\hline
          & \tabH ent & 73.51 & 62.79 & 69.88 & 61.96 & 70.85 & 72.56 \\
      BERT& ent+con & \textbf{77.79} & \textbf{71.11} & \textbf{78.31} & \textbf{75.23} & \textbf{81.24} & \textbf{83.73} \\
    -large& ent+rev & 69.46 & 54.67 & 66.23 & 55.28 & 68.13 & 64.73 \\
          & ent+con+rev & 77.02 & 68.02 & \textbf{77.86} & 73.65 & \textbf{83.73} & 80.99 \\
\hline
          & \tabH ent & 72.10 & 59.98 & 68.77 & 58.39 &  67.50 & 67.02 \\
   RoBERTa& ent+con & \textbf{77.60} & \textbf{70.58} & \textbf{78.76} & \textbf{75.90} & \textbf{81.21} & \textbf{83.26} \\
     -base& ent+rev & 69.35 & 54.21 & 66.19 & 54.50 & 66.54 & 61.90 \\
          & ent+con+rev & 76.37 & 66.39 & \textbf{77.74} & 73.01 & \textbf{82.95} & 80.46 \\
\hline
          & \tabH ent & 72.54 & 60.74 & 68.67 & 60.21 & 69.96 & 72.01 \\
   RoBERTa& ent+con & \textbf{78.05} & \textbf{71.41} & \textbf{79.96} & \textbf{78.12} & 81.05 & \textbf{84.91} \\
    -large& ent+rev & 69.17 & 54.54 & 66.47 & 55.96 & 67.84 & 68.05\\
          & ent+con+rev & 76.68 & 67.14 & 79.07 & 75.58 & \textbf{84.17} & 82.41 \\
\hline
\end{tabular}
\caption{Experimental results of the NLI task for all combination of a pre-trained model and loss function.
}
\label{tab:nli_full}
\end{table*}

Table~\ref{tab:nli_full} shows experimental results of the NLI task for all pre-trained models.
In addition to accuracy (Acc.), we adopted area under the precision-recall curve (AUPRC) as the evaluation metrics for this NLI task. 
To calculate the AUPRC, we varied the threshold for determining whether two sentences were in an entailment relation from 0 to 1 in steps of 0.001.

\section{Full Results of the Entailment Direction Prediction Task}
\label{sec:ent_full}

Table~\ref{tab:ent_full} shows experimental results of the entailment direction prediction task for all combinations of pre-trained models and loss functions.

\section{Detail of Experimental Setup}
\label{sec:detail}
The fine-tuning epoch size is 3, the temperature hyperparameter is 0.05, and the optimizer is AdamW~\cite{AdamW}.
The embedding dimensions were 768 for BERT-base and RoBERTa-base and 1024 for BERT-large and RoBERTa-large.
These settings are the same as SimCSE~\cite{SimCSE}.
Fine-tuning for BERT-base and RoBERTa-base took about 40 minutes on a single NVIDIA A100. 
Fine-tuning for BERT-large and RoBERTa-large took about 2 hours on the same GPU.
We carry out grid-search of batch size $\in \{16, 32, 64, 128\}$ and learning rate $\in \{1\mathrm{e}{-5}, 3\mathrm{e}{-5}, 5\mathrm{e}{-5}\}$ on the SNLI development set, then used the best-performing combination in the in-training evaluation described below.
The learning rate is 0 at the beginning and increases linearly to a set value in the final step.
Table~\ref{tab:grid} summarizes the detailed grid-search results. The values in the table represent the AUC values of the precision-recall curve for the NLI task for each batch size and learning rate, where each value was multiplied by 100.

In each experiment, the AUC of the precision-recall curve for the NLI task on the SNLI development set was calculated every 100 training steps, and the model with the best performance was used for the final evaluation on the test set. 

\begin{table}[t]
\centering
\small
\begin{tabular}{@{ \ }l@{ \ }l|c@{ \ \ }c@{ \ \ }c|c@{ \ }}
\hline
\tabH Model & Loss function & SNLI & MNLI & SICK & Avg.\\
\hline
\multicolumn{2}{@{ }l|}{\tabH Length-baseline} & 92.63 & 82.64 & 69.14 & 81.47\\
\hline
          &\tabH ent & 64.84 & 61.11 & 60.10 & 62.01\\
    BERT  & ent+con & 64.55 & 56.84 & 69.67 & 63.68\\
    -base & ent+rev & \textbf{97.60} & \textbf{92.64} & \textbf{87.80} & \textbf{92.68}\\
          & ent+con+rev & \textbf{97.38} & \textbf{91.92} & \textbf{86.22} & 91.84\\
\hline
          & \tabH ent & 62.06 & 60.09 & 62.09 & 61.41\\
      BERT& ent+con & 62.43 & 54.87 & 69.01 & 62.10\\
    -large& ent+rev & \textbf{97.66} & \textbf{92.76} & \textbf{88.03} &   \textbf{92.81}\\
          & ent+con+rev & \textbf{97.55} & \textbf{93.11} & \textbf{85.94} & 92.20\\
\hline 
          & \tabH ent & 65.84 & 60.41 & 59.69 & 61.98\\
   RoBERTa& ent+con & 65.66 & 55.24 & 69.97 & 63.62\\
     -base& ent+rev & \textbf{97.74} & \textbf{93.15} & \textbf{87.90} &         92.93\\
          & ent+con+rev & \textbf{97.44} & \textbf{93.10} & \textbf{88.43} & \textbf{92.99}\\
\hline
            & \tabH ent & 66.91 & 60.88 & 61.56 & 63.11\\
    RoBERTa & ent+con & 64.57 & 55.31 & 71.38 & 63.75\\
    -large  & ent+rev & \textbf{97.89} & \textbf{93.97} & \textbf{88.71} & \textbf{93.52}\\
            & ent+con+rev & \textbf{97.42} & 93.61 & 86.57 & 92.53\\
\hline
\end{tabular}
\caption{Experimental results of the entailment direction prediction task for all combinations of pre-trained models and loss functions.
}
\label{tab:ent_full}
\end{table}

\begin{table}[h]
\centering
\small
\begin{tabular}{lc|ccc}
\hline
\multirow{2}{*}{Model} & \multirow{2}{*}{Batch size} & \multicolumn{3}{c}{\tabH Learning rate}\\
 & & 1e-5 & 3e-5 & 5e-5\\
\hline
\multirow{4}{*}{BERT-base} & \tabH 16 & 63.05 & 65.72 & \textbf{66.21}\\
\         & 32 & 62.02 & 64.69 & 64.84\\
\         & 64 & 60.44 & 62.93 & 64.20\\
\         & 128 & 58.99 & 61.26 & 62.66\\
\hline
\multirow{4}{*}{BERT-large} & \tabH 16 & 64.66 & \textbf{65.65} & 61.09\\
\          & 32 & 63.73 & 65.56 & 63.42\\
\          & 64 & 62.24 & 65.01 & 62.46\\
\          & 128 & 60.72 & 63.41 & 64.68\\
\hline
\multirow{4}{*}{RoBERTa-base} & \tabH 16 & 64.66 & 65.78 & \textbf{66.31}\\
\            & 32 & 63.06 & 65.09 & 65.68\\
\            & 64 & 61.59 & 64.18 & 64.95\\
\            & 128 & 60.48 & 62.54 & 63.84\\
\hline
\multirow{4}{*}{RoBERTa-large} & \tabH 16 & 66.22 & \textbf{67.17} & 61.69\\
\             & 32 & 65.96 & 67.10 & 60.64\\
\             & 64 & 64.26 & 66.01 & 66.88\\
\             & 128 & 63.07 & 64.91 & 65.72\\
\hline
\end{tabular}
\caption{Grid-search results.}
\label{tab:grid}
\end{table}

\section{Ratio of Length of Sentence Pairs}
\label{sec:ratio}

\begin{figure}[!t]
\begin{center}
 \includegraphics[clip, width=1.0\linewidth]{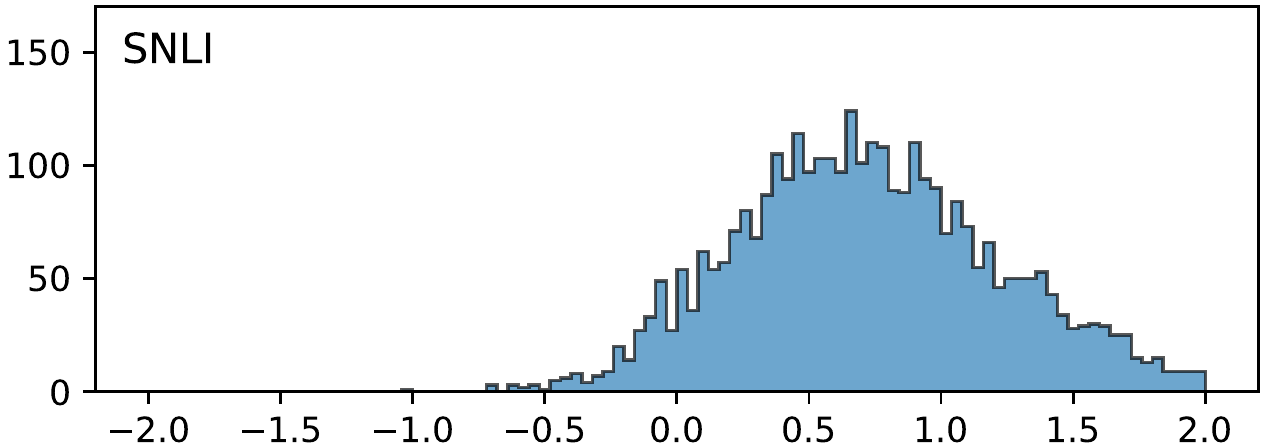}
\includegraphics[clip, width=1.0\linewidth]{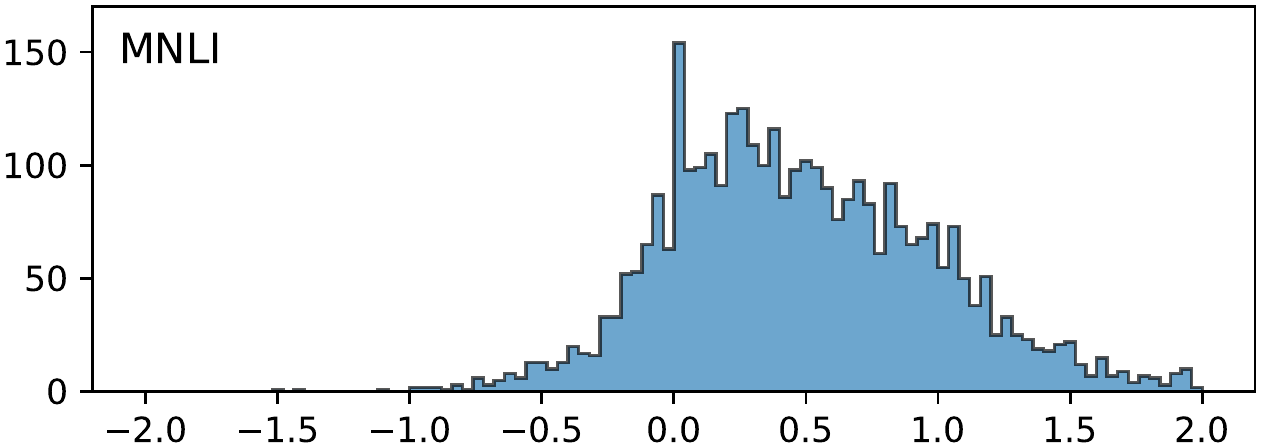}
 \includegraphics[clip, width=1.0\linewidth]{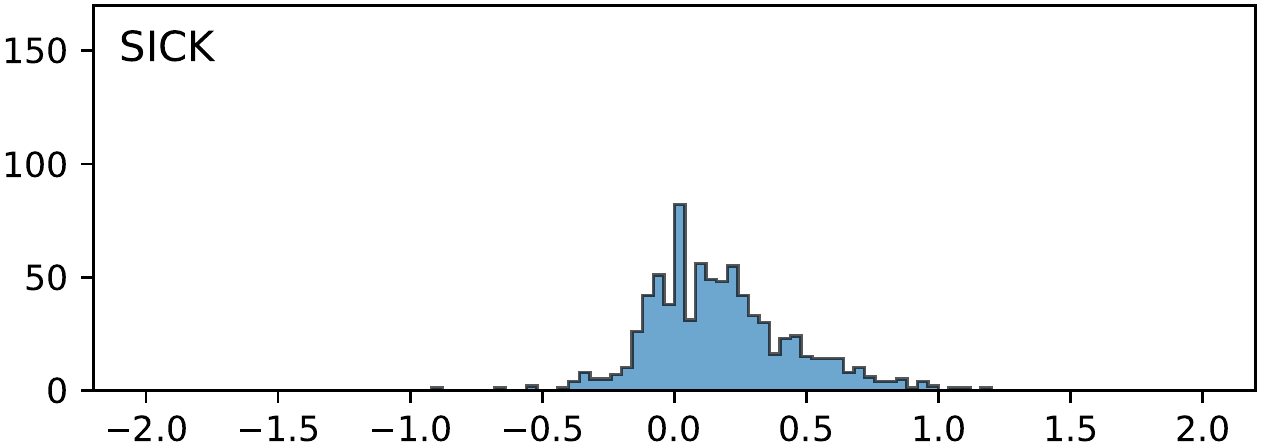}
\end{center}
\caption{Histograms representing the distributions of the logarithmic values of the length ratios of the premise sentences and their corresponding hypothesis sentences in the SNLI, MNLI, and SICK datasets. The horizontal axis represents the logarithm of the length ratio, and the vertical axis represents the number of sentence pairs.}
\label{fig:len}
\end{figure}

Figure~\ref{fig:len} shows histograms of the ratios of the length of the premise sentence to that of the hypothesis sentence for each sentence pair in each dataset.

\end{document}